\documentclass[letterpaper]{article} % DO NOT CHANGE THIS
\usepackage{aaai24}  % DO NOT CHANGE THIS
\usepackage{times}  % DO NOT CHANGE THIS
\usepackage{helvet}  % DO NOT CHANGE THIS
\usepackage{courier}  % DO NOT CHANGE THIS
\usepackage[hyphens]{url}  % DO NOT CHANGE THIS
\usepackage{graphicx} % DO NOT CHANGE THIS
\usepackage{natbib}  % DO NOT CHANGE THIS AND DO NOT ADD ANY OPTIONS TO IT
\usepackage{caption} % DO NOT CHANGE THIS AND DO NOT ADD ANY OPTIONS TO IT
\usepackage{algorithm}
\usepackage{algorithmic}
\usepackage{amsmath}
\usepackage{xcolor}
\usepackage{mathtools}
\usepackage{amsfonts}

\usepackage{inconsolata}
\usepackage{graphicx}
\usepackage{multirow}
\usepackage{newfloat}
\usepackage{listings}
\DeclareCaptionStyle{ruled}{labelfont=normalfont,labelsep=colon,strut=off} % DO NOT CHANGE THIS
\floatstyle{ruled}
\newfloat{listing}{tb}{lst}{}
\floatname{listing}{Listing}
\title{SpikingBERT: Distilling BERT to Train Spiking Language Models Using Implicit Differentiation}
\author{
    Malyaban Bal,
    Abhronil Sengupta
}
\affiliations{
    School of Electrical Engineering and Computer Science\\
  The Pennsylvania State University\\
  University Park, PA 16802 \\
    mjb7906@psu.edu, sengupta@psu.edu
}

\usepackage{bibentry}
\begin{document}

\maketitle

\begin{abstract}
Large language Models (LLMs), though growing exceedingly powerful, comprises of orders of magnitude less neurons and synapses than the human brain. However, it requires significantly more power/energy to operate. In this work, we propose a novel bio-inspired spiking language model (LM) which aims to reduce the computational cost of conventional LMs by drawing motivation from the synaptic information flow in the brain. In this paper, we demonstrate a framework that leverages the average spiking rate of neurons at equilibrium to train a neuromorphic spiking LM using implicit differentiation technique, thereby overcoming the non-differentiability problem of spiking neural network (SNN) based algorithms without using any type of surrogate gradient. The steady-state convergence of the spiking neurons also allows us to design a spiking attention mechanism, which is critical in developing a scalable spiking LM. Moreover, the convergence of average spiking rate of neurons at equilibrium is utilized to develop a novel ANN-SNN knowledge distillation based technique wherein we use a pre-trained BERT model as ``teacher'' to train our ``student'' spiking architecture. While the primary architecture proposed in this paper is motivated by BERT, the technique can be potentially extended to different kinds of LLMs. Our work is the first one to demonstrate the performance of an operational spiking LM architecture on multiple different tasks in the GLUE benchmark. Our implementation source code is available at https://github.com/NeuroCompLab-psu/SpikingBERT.
\end{abstract}

\section{Introduction}

Large language Models (LLMs) are becoming increasingly popular because of its broad applications in a variety of natural language processing (NLP) tasks. LLMs like GPT-3 \cite{brown2020language} has shown additional characteristics such as emergent abilities \cite{wei2022emergent} which can only be realized once the model size/compute increases above a certain threshold. Recent times have witnessed commercial deployment of LLMs enabling worldwide reach and positively impacting real-world users.
However, the immense power of LLMs comes at the cost of huge energy consumption both during the computationally expensive training phase as well as the inference phase. LLMs are characterized by large number of trainable parameters and are usually very deep. In order to alleviate the operational complexity of LLMs, we aim to draw motivation from the brain. Integrating the mechanism and knowledge embodied in LLMs into brain-inspired neural models hold immense promise for creating a bio-plausible and energy-efficient solution.

Spiking neural networks (SNNs) \cite{ghosh2009spiking} are biologically inspired and communication between two neurons in an SNN architecture occurs in the form of spikes. This sparse spike-based information flow enables event-driven computation and communication in neuromorphic hardware, thereby resulting in significant energy savings \cite{sengupta2019going}. SNN based architectures have been also tested extensively on neuromorphic hardware like Intel's Loihi 2 processor  \cite{davies2021advancing} and have demonstrated orders of magnitude energy efficiency. 

% SNNs have demonstrated performance close to conventional ANN neural architectures primarily on vision based datasets \cite{sengupta2019going,schuman2022opportunities}. Preliminary work has recently explored their applicability on NLP tasks \cite{mueller2021spiking}. 

% There have been popular hypothesis that neurons collectively settle to different configurations \cite{hinton2002training} according to the sensory inputs. This allows for efficient processing of the input for any further output. 

Scaling SNNs to complex domains, like NLP, poses significant challenges mainly due to the absence of scalable and efficient learning algorithms. The complexity of the tasks, coupled with the growing depth of the required model architectures, renders the practical application of BPTT infeasible. In this work, backed by robust theoretical foundations and empirical evidence, we explore a scalable framework for training spiking LMs. We consider our spiking LM as a dynamical system that, given an input, progressively converges to a steady-state (over $T_{conv}$ time steps). Similar to most supervised learning algorithms, training is done in two phases, viz. ``forward'' and ``backward''. However, instead of learning through unrolling the computational graph over the operated time steps (like in BPTT), we leverage the convergence of the average spiking rate (ASR) of the neurons to an equilibrium state during the ``forward'' phase of learning. Upon convergence, we can derive fixed-point equations from the underlying model and subsequently employ implicit differentiation on the attained steady-state to train the model parameters effectively as described later on. 

Training using implicit differentiation is primarily used in deep equilibrium models (DEQ) \cite{bai2019deep}. Recently, this method has also been used for training of convolution based spiking architectures \cite{xiao2021training} for vision related tasks. This methodology offers exceptional memory efficiency during training unlike BPTT, which requires a huge amount of memory to store a large computational graph. It also eliminates the necessity of surrogate gradient methods by implicitly calculating gradients, thereby circumventing the non-differentiability problem of spiking models. Under certain constraints \cite{bai2019deep}, this form of learning is similar to bio-plausible and energy-based training methods like equilibrium propagation \cite{scellier2017equilibrium, bal2022sequence}, thus bolstering a neuromorphic viewpoint of learning.

In transformer \cite{vaswani2017attention} based LMs as discussed in this paper, the attention mechanism serves as a vital component. However, vanilla attention mechanism is fundamentally non-spiking in nature, as it relies on sequences of real-valued vectors for the Query, Key, and Value components.  In this paper, we present a spiking attention mechanism that utilizes spike-based inputs and operates over the number of time steps ($T_{conv}$) required for model convergence. The convergence of ASR of the neurons at equilibrium allows us to draw a close equivalence between the ASR of the spiking attention layer and vanilla attention. 

Training LMs from scratch is a significant time and resource-intensive process. The additional overheads of training a spiking LM from scratch prompted us to seek out more proficient approaches for training our model. Knowledge distillation (KD) \cite{hinton2015distilling} allows for faster and efficient transfer of knowledge from a trained ``teacher'' model to a possibly smaller in size ``student'' model. In this paper, we leverage the steady state ASR of the spiking LM and propose a novel ANN-SNN KD framework involving the ASR at equilibrium of specific intermediate layers of the ``student'' model and the activation of target layers of a larger pre-trained ``teacher'' model. Moreover, the feasibility of model training during distillation is enabled by the previously discussed training method. 
The primary contributions of our work are as follows:
\begin{itemize}
\item \noindent \textbf{SpikingBERT with Spiking Attention}: We propose a fully-operational spiking LM, following the architecture of BERT \cite{devlin2018bert}, and evaluate it against different tasks (classification and regression) of the GLUE benchmark \cite{wang2018glue}. We also propose an efficient spiking attention mechanism whose ASR at equilibrium approximates vanilla non-spiking attention. 
\item \noindent \textbf{Spiking LM Training: }We theoretically and empirically verify the convergence of our proposed spiking LM (comprising both linear and non-linear operations) to equilibrium state and use an implicit differentiation based method to overcome the non-differentiability issue of SNN training and reduce memory usage during training. This method enables training of Spiking LMs that surpass the scale of existing spiking models, thereby allowing development of deeper models for complex tasks.
\item \noindent \textbf{ANN-SNN KD using Equilibrium States: } We leverage the equilibrium state of the neurons after convergence, to train our model more effectively using a novel ANN-SNN KD framework. This allows for developing an efficient and smaller spiking ``student'' model using larger BERT models as its ``teacher''.
\end{itemize}

\section{Related Works}
\textbf{Spiking Architectures:} Spiking architectures \cite{sengupta2019going, lee2020enabling, xiao2021training,  zhou2022spikformer} have primarily been explored for vision based datasets such as CIFAR-100 \cite{krizhevsky2009cifar}, ImageNet \cite{deng2009imagenet}, among others as well as neuromorphic datasets such as NMNIST \cite{orchard2015converting}, DVS Gesture \cite{amir2017low}. However, limited work has been done for sequence based NLP tasks. While most of the networks are non-attention based shallow models \cite{alawad2017energy}, recently \cite{zhu2023spikegpt} explored developing a GPT-like model using linear attention.
While GPT \cite{radford2018improving} is a decoder-only architecture, BERT is an encoder-only architecture whose ability to capture bi-directional contextual information makes it more suitable for text classification problems. Unlike our SpikingBERT, the SpikeGPT model uses surrogate gradient to overcome the non-differentiability problem and uses a  BPTT approach for training. Furthermore, SpikingBERT is the first spiking LM, to the best of our knowledge, that has been evaluated against multiple different tasks in the GLUE benchmark. Moreover, due to the efficient KD incorporated in our approach, we can enhance model performance without the need for an extensive number of parameters. There has been some work on KD in SNNs previously, however all of them primarily explored BPTT based methods and focused solely on simple vision based datasets \cite{xu2023constructing, takuya2021training, hong2023lasnn}. 

\noindent \textbf{Efficient LMs}: Given the increasingly growing scale of LMs, research focusing on attempting to make them computationally less expensive and smaller in size have gained significant attention. TinyBERT \cite{jiao2019tinybert} proposed extracting knowledge from the ``teacher" model - from both the intermediate layers and the prediction layer. NAS-BERT \cite{xu2021bert} does model compression using neural architecture search. Work has also been done on distilling knowledge from BERT to a single layer BiLSTM network \cite{tang2019distilling}. Moreover, research endeavours have been made in methods like quantization \cite{kim2021bert}, pruning \cite{kurtic2022optimal}, etc. to reduce the model complexity. Because appropriate neuromorphic baselines are currently unavailable, we compare our proposed model with existing standard NLP models and efficient LMs. 

% Nevertheless, our approach stands apart fundamentally: its operations are aligned with bio-plausible principles, allowing deployment on neuromorphic chips to harness its energy-saving advantages.

\section{Methods}

In this section, we will begin by examining the foundational principles of our method. Subsequently, we will delve into the architectural details to provide a comprehensive understanding. We delve into the theoretical and empirical foundations underlying the convergence of ASR in SNNs trained using implicit differentiation. Furthermore, we present an innovative approach to harness this convergence for the design of a spiking attention mechanism and a novel KD mechanism leveraging a pre-trained BERT model as a ``teacher'', thereby enhancing the learning process. We also elaborate on the framework for using implicit differentiation based technique to train our spiking architecture.

\subsection{Spiking Neural Networks}
The fundamental building block of the proposed spiking architecture comprises of leaky integrate and fire (LIF) neurons. The internal dynamics of a simple LIF neuron is similar to a biological neuron and is given as follows, 

\begin{equation}
\label{eqn1}
{\tau}_{m} \cdot \frac{du}{dt} = - (u(t) - u_{rest}) + R \cdot I(t) 
\end{equation}
where, $u$ is the membrane potential; $u_{rest}$ is the resting membrane potential;  $I(t)$ is the input voltage at time $t$ scaled by a constant, $R$, representing resistance; ${\tau}_{m}$ is the time constant. Moreover, if $u > V_{th}$, then the neuron potential is updated by subtracting $V_{th}$ and the neuron emits a spike, i.e., it sends out a `1' else `0'. A suitable discrete time representation of the dynamics for the $i^{th}$ neuron, can be described as follows,

\begin{equation}
\label{eqn2}
\begin{aligned}
u_i[t + \delta] = \gamma  u_i[t] + \sum_j(w_{ij}s_j[t]) + b_i, \\
s_i[t+1] = S(u_i[t + \delta]), \\
u_i[t + 1] = u_i[t + \delta] - V_{th} s_i[t+1]
\end{aligned}
\end{equation}
where, $\gamma$ is the leaky term related to the constant ${\tau}_{m}$ in Eqn. \ref{eqn1} (for LIF neurons, we keep $\gamma < 1$ and for IF $\gamma = 1$); $s_j$ is the spike from the $j^{th}$ input neuron; $w_{ij}$ is the synaptic weight of the connection between the pre-synaptic and post-synaptic neurons; $t+\delta$ represents an intermediate time step representation to determine if the neuron has fired; $b_i$ represents a bias term; $S$ is the non-differentiable function for spike generation and subtraction is used as reset operation.

% Thus, communication between two neurons occurs through binary spikes resulting in conditional event-driven computation and data communication.

\begin{figure}
  \centering
  \includegraphics[width=1\columnwidth]{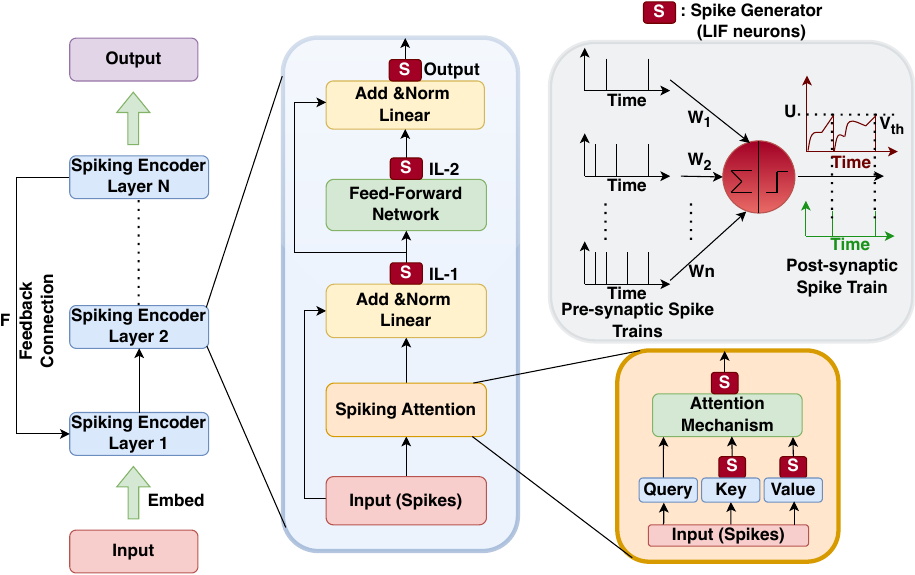}
  \caption{High-level overview of the SpikingBERT model. During the ``forward'' phase of learning, the network is simulated over $T_{conv}$ time steps, i.e., until the ASR of the neurons of each layer converges to an equilibrium. Information flow both within and between two spiking encoders occur using spikes instead of real values, thereby mimicking event-driven information flow in bio-inspired systems.}
  \label{fig1}
\end{figure}

\subsection{Implicit Modeling}
Implicit modeling takes a different approach by not explicitly defining the precise computation of a model's output from its input. Instead, it relies on imposing specific constraints on the model, ensuring that these constraints are met to achieve the desired results. For example,  consider a simple model represented by a function $h$. In order to formulate an explicit model with input $x \in X$ and output $z \in Z$, the following computation is performed: $ z = h(x)$. However, for formulating it implicitly, a function $g: X \times Z \rightarrow R^n$ is defined, such that $g(x,z) = h(x) - z$ and the goal will be to find the root of the equation: $g(x,z) = 0$.
While this simple example demonstrates algebraic equations, these methodologies can be extended to fixed-point equations, thereby paving the way for the development of DEQ \cite{bai2019deep}.

Let us consider a fixed-point equation of the form $z = f_{\theta}(z)$, where $\theta$ is the set of parameters. This fixed point equation converges over time, i.e., after $T_{conv}$ time steps $z_{T_{conv}} = z_{(T_{conv}+1)}$, thereby reaching an equilibrium state. Similarly, as before, we can form another equation namely, $g_{\theta}(z) = f_{\theta}(z) - z$. Here, the loss function $L$ that we will be defining will utilize the value of $z$ at equilibrium, i.e., $z_{T_{conv}} = z^{*}$. Using implicit differentiation \cite{bai2019deep}, the following relation can be derived,

\begin{equation}
\label{eqn3}
\frac{\partial L(z^*)}{\partial \theta} = - \frac{\partial L(z^*)}{\partial z^*} (J^{-1}_{g_{\theta}}|_{z^*}) \frac{\partial f_{\theta}(z^*)}{\partial \theta}
\end{equation}
where, $J^{-1}_{g_{\theta}}|_{z^*}$ is the inverse Jacobian of $g_\theta$ when $z = z^{*}$, i.e., at equilibrium. The proposed spiking LM architecture follows a similar set of equations which will be described in the following section. Since the gradient is computed using implicit differentiation on the converged steady-state, we avoid the non-differentiability issues of the spiking function \cite{neftci2019surrogate}. Furthermore, by computing gradients solely at the equilibrium state, there is no requirement to store intermediate hidden states. This characteristic enhances the scalability and memory efficiency of this approach in comparison to BPTT.

\subsection{Architecture}

% \textcolor{blue}{The proposed architecture is primarily based designed using spiking IF/LIF neurons.} 

The high-level building block of the proposed spiking LM comprises of Spiking Encoder (SE) layers, which can be considered similar to individual encoder layers in a transformer architecture. Both intra and inter-layer communication in the SE layers occur using spikes at every time step during the ``forward'' phase and spiking LIF neurons are fundamental units in its design. As shown in Fig. \ref{fig1}, each SE layer consists of a Spiking Attention layer followed by fully connected layers (some including skip-connections similar to those in BERT) viz. Intermediate Layer-1 (IL-1), Intermediate Layer-2 (IL-2) and an output layer, all of which operate using spikes. The structure of the proposed network comprises of $N$ stacked SE layers similar to the structure of BERT. The input embeddings are processed by an LIF neuron layer, generating spikes that are propagated through the model. In some of the internal layers of SE, we use layer normalization (following BERT). Effect of normalization and other architectural details are added in Appendices B \& C.

% \textcolor{red}{While this enables faster convergence, the proposed model can still operate without it with similar performance as is shown in the technical appendix.}

From a biological perspective, feedback connections are present in the human-brain and moreover in some cases \cite{kubilius2019brain} shallower network with recurrent connections shows performance comparable or better than deeper architectures. The connection ($F$) is added from the output of the final SE layer to the first one in order to introduce a feedback. The feedback connection is an optional component that adds to the model's bio-plausibility. The general formulations of steady-state ASR equations, developed subsequently, can be seamlessly applied to models involving both feedback connections and those without any feedback. Unlike in vision-based tasks, feedback did not improve performance considerably when compared with no-feedback scenario in our experiments with GLUE benchmark. However, we still explore it on a theoretical level to maintain consistency with previous works \cite{xiao2021training} and to encourage future research on feedback enabled SNNs in other domains.

Similar to vanilla transformer based architectures, the input sequence is directly fed into the model. The model converges over $T_{conv}$ time steps during the ``forward'' phase to settle to an equilibrium state. As discussed earlier, the spiking neurons of the model have their individual membrane potentials, $u$, which are updated at every time step during convergence.  At time $t+1$, the membrane potential of $u_1$, i.e., the input to the first SE layer can be given as,
\begin{equation}
\begin{aligned}
\label{eqn4}
u_{1}[t+1] = \gamma u_{1}[t] + F s_{(N,out)}[t]  \\ + W_0(x)
+ b_{1} - V_{th}s_{1}[t+1]
\end{aligned}
\end{equation}
where, $W_0$ provides the embedding of the input sequence $x$ of length $N_s$ and produces a sequence of vectors $y \in \mathbb{R}^{N_s \times D_{emb}}$ ($D_{emb}$ is the encoding dimension), $F$ is the weight of the feedback connection (if feedback is included), $b_1$ is bias and $s_{(N,out)}[t]$ are spikes generated from the $N^{th}$ SE layer in the previous time step.
The membrane potential of a layer $i>1$ can be represented simplistically as,
\begin{equation}
\begin{aligned}
\label{eqn5}
u_{i}[t+1] = \gamma u_{i}[t] + W_{(i-1)}(s_{(i-1)}[t+1])  + b_{i}  \\- V_{th}s_{i}[t+1]
\end{aligned}
\end{equation}
where, $W_{(i-1)}$ is an operation (as formulated by each individual layer) defined on a set of spikes from previous layers. The described LIF neurons propagate information using spikes that are generated following Eqn. \ref{eqn2}.

The average spiking rate (ASR) of a neuron at layer $i$ can be defined as, $a_i[t] =  \frac{\sum^t_{\tau = 1}\gamma^{t-\tau}s_i[\tau]}{\sum^t_{\tau = 1}\gamma^{t-\tau}}$. Given $W_{(i-1)}$ is a linear operation, using Eqn. \ref{eqn5} and performing a weighted (if $\gamma < 1$) average over time (with $u[0]=0, s[0]=0$) we get,

\begin{equation}
\label{eqn6}
\begin{aligned}
a_{i}[t+1] = \frac{1}{V_{th}}(W_{(i-1)}a_{(i-1)}[t+1] + b_i - \frac{u_{i}[t+1]}{\sum^t_{i=0}\gamma^i})
\end{aligned}
\end{equation}

Since, $a_i[t]$ represents ASR, its value is restricted within [0,1]. Following previous work on implicit differentiation at equilibrium \cite{xiao2021training}, 
% we can define $u_i[t] = u^-_i[t] + u^+_i[t]$, where $\frac{1}{t}u^-_i[t] = min(max(v_i[t]-V_{th},0), v_t[t])$. Here, $v_i[t] = W_{(i-1)}a_{(i-1)}[t] + b_i$  and $u^+_i[t]$ is a bounded positive term. 
as the average of input converges to equilibrium $\Bar{x}[t] \rightarrow x^*$, then the ASR of the layers (Eqn. \ref{eqn6}) in the spiking architecture converges to equilibrium points: $a_i[t] \rightarrow a_i^*$ (with bounded random error in case of LIF neurons). At equilibrium, the ASR $a_i^*$ of layer $i$ satisfies,

\begin{equation}
\begin{aligned}
\label{eqn7}
a_i^* = \sigma (\frac{1}{V_{th}}(W_{(i-1)}(a^*_{i-1}) + b_i))
\end{aligned}
\end{equation}
where clipping function $\sigma(x)$ bounds the values within [0,1].

\begin{figure}
  \centering
  \includegraphics[width=\columnwidth]{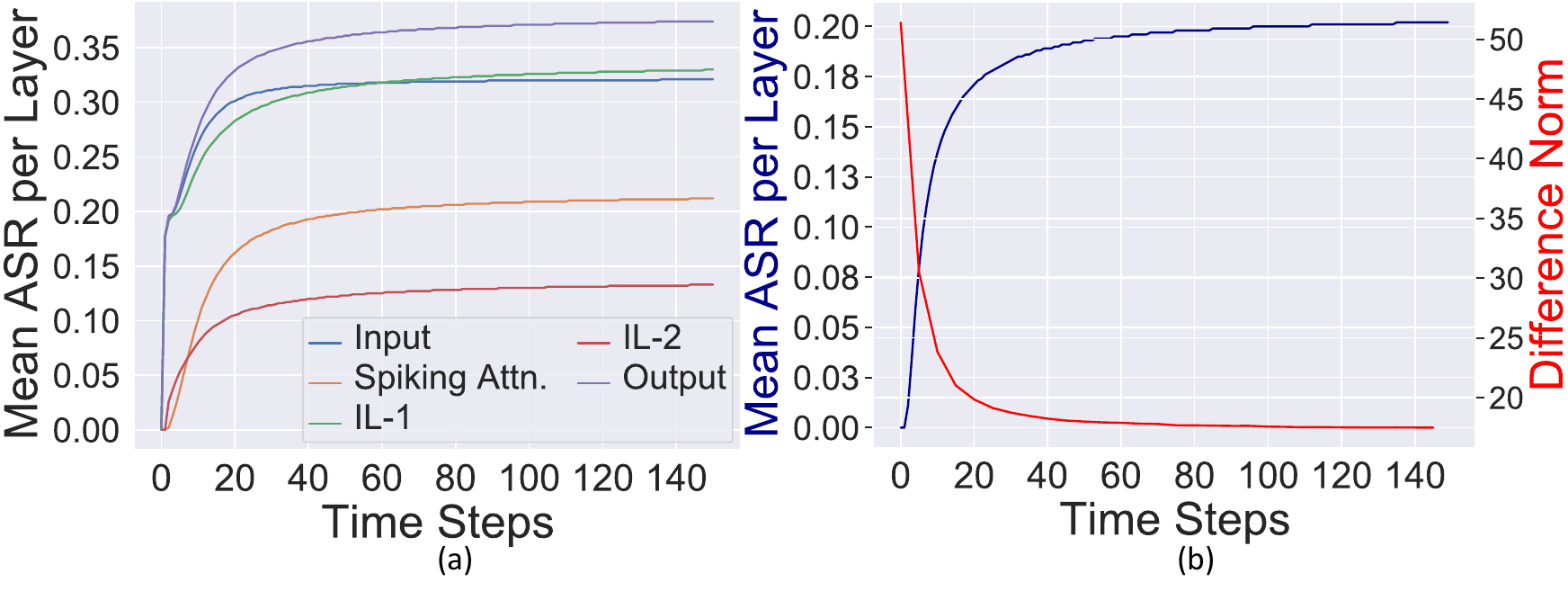}
  \caption{Results obtained after passing a randomly sampled input from SST-2 dataset through SpikingBERT\textsubscript{4}. (a) Graph showing mean (over number of neurons) of the ASR of different sub-layers in an SE layer against the operating time steps. (b) The y-axis on the left depicts mean (over number of neurons) of the ASR of a randomly chosen spiking attention layer. Along the right y-axis, the ``Difference Norm'' between the output of the steady-state equation of the chosen spiking attention layer and the calculated ASR is shown. Time steps used for convergence in shown along the x-axis.}
  \label{conv_fig}
\end{figure} 
Like the linear operations, the layers with non-linear operations such as spiking attention also converges to a steady-state ASR as is empirically validated in this paper (Fig. \ref{conv_fig}a). Steady-state equations like Eqn. \ref{eqn7} are leveraged during training. Thus, for the spiking attention layer, we formulate a surrogate steady state equation at equilibrium given as,

\begin{equation}
\begin{aligned}
\label{eqn8}
a_{(attn)}^* = \sigma (\frac{1}{V_{th}}(Attn(a^*_{x}, a^*_{k}, a^*_{v}) + b_{(attn)})
\end{aligned}
\end{equation}
where, $a_x^*$ is the ASR of the layer used to form Query, $a_k^*$ is the ASR of the Key and $a_v^*$ is the ASR of the Value. Operational details of the spiking attention mechanism and empirical convergence and justification of the defined equation is discussed in the next subsection. 

Thus, after the model converges to equilibrium, the dynamics of the steady-state ASR of the underlying SNN can be mapped to a surrogate non-spiking architecture where the input and output of each layer are the corresponding ASRs ($a^*_i$). The operation of individual layers in the surrogate network is given by the steady-state equations as described earlier and  can be simplified to the form,
$a_{i}^* = l_{i}(a_j^*, \dots)$ where,  $l_{i}$s are steady-state equations corresponding to each layer like Eqn. \ref{eqn7} and \ref{eqn8}. The parameters ($a^*_j, \dots$) associated with each layer ($l_{i}$) are defined according to the specific operation. If we use feedback connection, the fixed-point equation of the first layer is of the form $a_1^* = l_1(l_M \circ \dots \circ l_2(a_1^*), x^*)$ where, $l_1(a,x) = \sigma (\frac{1}{V_{th}}(Fa + W_0(x)+b_1))$ with $M$ being the total number of individual layers.

 For the task of text classification, we use the last layer of the network, i.e., the output of the $N^{th}$ encoder layer given as $a^*_{N, out}$ as an input to a linear classification function. Simulating the network for $T_{conv}$ time steps, we can compute $a_{N, out}[T] = \frac{\sum_t(s_{N, out}[t])}{T}$ (for simplicity of demonstration, $\gamma = 1$), which we can use as $a^*_{N, out}$.
 Moreover, since the behaviour at equilibrium is captured by the surrogate network using only ASR, we can simply perform backpropagation to train the weights by leveraging Eqn. \ref{eqn3}. Thus, instead of performing BPTT to train the underlying spiking architecture, we use simple backpropagation to train the weights of the spiking LM using only equilibrium state ASR of neurons.

\subsection{Spiking Attention Mechanism}

We propose a computationally efficient Spiking Attention mechanism where the inputs are processed as spikes from the previous layer. The proposed attention operations of the module at time step $t$ can be formulated as,

\begin{equation}
\label{eqn9}
\begin{aligned}
Attn(S_x(t), S_K(t), S_V(t)) = \\ \pi(s * Q(S_x(t)) (S_K(t))^T)\cdot S_V(t)
\end{aligned}
\end{equation}
where, $Q(S_x(t))$ is obtained after passing input spikes ($S_x(t)$) at time $t$ through a linear layer ($W_Q$) for generating Query. The spikes corresponding to the Key layer ($S_K(t)$) is computed by passing the input spikes $S_x(t)$ through a linear mapping ($W_K$), connected to an LIF neuron layer, as illustrated in Fig. \ref{fig1}. $S_V(t)$ is obtained similarly using linear mapping ($W_V$). $\pi$ is usually the softmax function and $s$ is a scaling factor where generally $s = \frac{1}{\sqrt{d_k}}$, with $d_k$ being the encoding dimension of Key. Recent work has shown that the non-linear normalization operation $\pi$ is not always essential \cite{zhou2022spikformer}. The operations outlined in Eqn. \ref{eqn9} exhibit characteristics akin to spiking architectures. This is because performing the aforementioned matrix multiplications entails multiplying a real-valued matrix with a matrix composed of spikes (i.e., `0's and `1's) at each step. Thus, instead of requiring $O(n^3)$ floating point multiplicative and $O(n^3)$ accumulative operations, we can implement the attention mechanism utilizing only $O(n^3)$ accumulative operations - which has been shown to significantly reduce computation cost in SNNs \cite{sengupta2019going} (note that this is a first order estimate ignoring memory transactions). The output of this module is passed through an LIF neuron, resulting in spikes that are fed to the next layer (Fig. \ref{fig1}).

The empirical convergence of the ASR of attention layer is demonstrated in Fig. \ref{conv_fig}b. As discussed earlier, we construct a surrogate steady-state function at equilibrium which helps us in efficient training of the model. The empirical rationale for employing this specific functional form is substantiated by observing the reduction in the difference norm between the output of the surrogate equation and the computed ASR of the layer at each timestep, as demonstrated in Fig. \ref{conv_fig}b. Thus, using Eqn. \ref{eqn8} and considering no bias, $V_{th} = 1$ and no clamping function $\sigma$, we see that as the model converges in time, the actual ASR of the spiking attention layer $a_{attn}[t]$ approximates vanilla attention given by $Attn(a_{x}[t], a_{k}[t], a_{v}[t])$.

\subsection{ANN-SNN KD using Equilibrium States}

\begin{figure}
  \centering
  \includegraphics[width=.9\columnwidth]{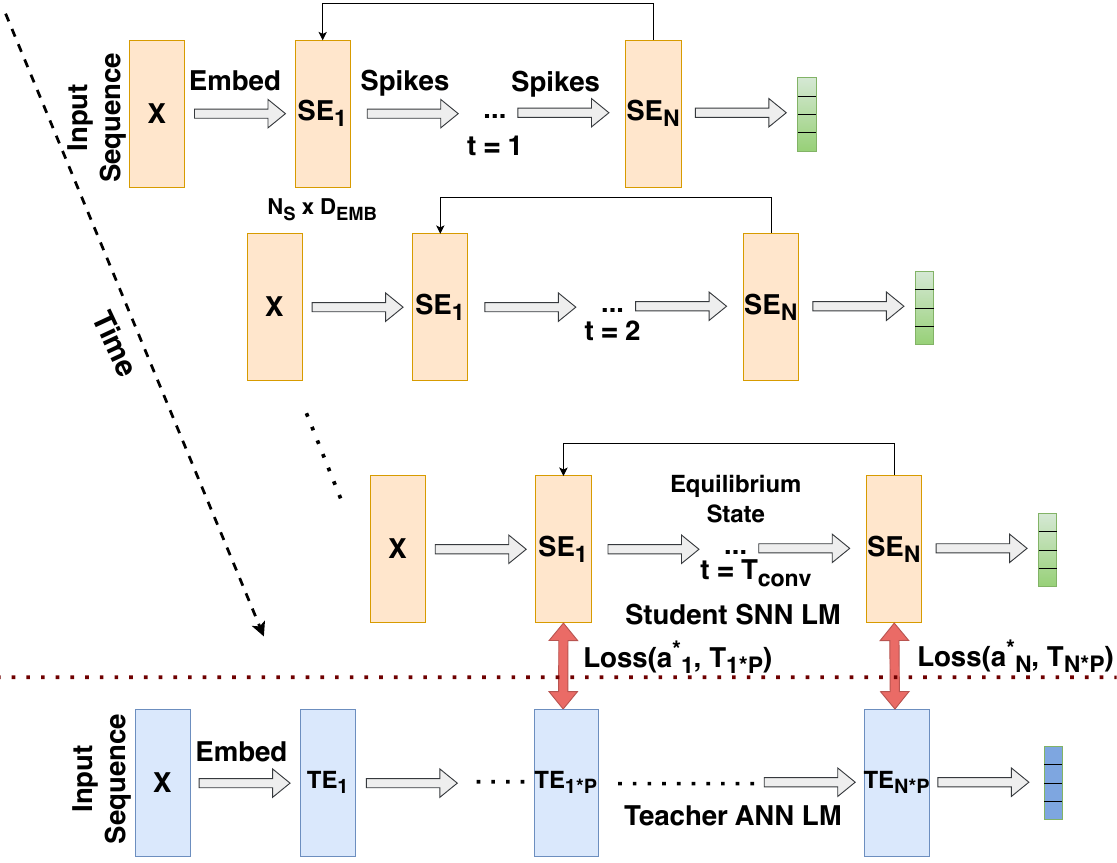}
  \caption{High-level overview of transformer layer based KD at equilibrium (following Eqn. \ref{eqn10}) from a ``teacher'' LM to a spiking ``student'' LM.}
  \label{KD_fig}
\end{figure}

\begin{table*}
\centering
\small
\begin{tabular}{llllllll}
\hline
\textbf{Model} & \textbf{QQP} & \textbf{MNLI-m} & \textbf{SST-2} & \textbf{QNLI} & \textbf{RTE} & \textbf{MRPC} & \textbf{STS-B} \\
\hline
CBoW \cite{wang2018glue} & 75.00 & 57.10 & 79.50 &  62.50 & 71.90 &  75.00/83.70 &70.60/71.10\\
BiLSTM \cite{wang2018glue} & 85.30 & 66.70 & 87.50 &  77.00 & 58.50 &  77.90/85.10 &71.60/72.00\\
BiLSTM + Attn, CoVe \cite{wang2018glue} & 83.50 & 67.90  & 89.20 &  72.50 & 58.10 & 72.80/82.40 & 59.40/58.00\\
GenSen \cite{wang2018glue} & 82.60 & 71.40  & 87.20 & 62.50 & 78.40 & 80.40/86.20 &81.30/81.80\\
BERT\textsubscript{5} + PF \cite{xu2021bert} & 84.10 & 67.70  & 81.60 & 80.90 & 62.80 & 78.60/- &-/81.10\\
NAS-BERT\textsubscript{5} + PF \cite{xu2021bert} & 85.70 & 74.20  & 84.90 & 83.90 & 67.00 & 80.00/- & -/82.80\\
NAS-BERT\textsubscript{5} + KD \cite{xu2021bert} & 85.80 & 74.40  & 87.30 & 84.90 & 66.60 & 79.60/- & -/83.00\\
% BERT\textsubscript{10} + KD \cite{xu2021bert} & 87.80 & 74.40  & 86.60 & 85.70 & 66.90 & 77.90/- &-/85.20\\
NAS-BERT\textsubscript{10} + PF \cite{xu2021bert} & 88.40 & 76.00  & 88.60 & 86.30 & 68.70 & 81.50/- &-/84.30\\
BERT\textsubscript{TINY} Adam \cite{frantar2021m} & 81.09 & 65.36  & 80.11 & 77.85 & - & 69.90/- &64.39/-\\
BERT\textsubscript{MINI} Adam \cite{frantar2021m} & 86.45 & 73.30 & 85.46 & 83.85 & - & 76.57/- &82.09/-\\
\textbf{SpikingBERT\textsubscript{4}} & \textbf{86.82} &  \textbf{78.10} &  \textbf{88.19} &  \textbf{85.20} &  \textbf{66.06} &  \textbf{79.17/85.15} &  \textbf{82.20/81.90}\\
\hline
TinyBERT\textsubscript{4} (no DA) \cite{jiao2019tinybert} & 88.50 & 80.60 & 90.50 & 87.00 & 68.20 & 82.40/- &86.20/85.70\\
\end{tabular}

\caption{
Results showing performance of our model (SpikingBERT\textsubscript{4}) against some standard models and other efficient implementations of BERT on GLUE evaluation set. Accuracy is used as the metric for QQP, MNLI-m, SST-2, QNLI, RTE datasets while both accuracy and F1 scores are reported for the MRPC dataset. For STS-B, we report Pearson/Spearman correlation. 
}
\label{table1}
\end{table*}

The proposed architecture and training mechanism guarantee the steady-state convergence of ASR of neurons across all layers in the models (Fig. \ref{conv_fig}a), including the internal representational layers. This enables us to develop an ANN-SNN based KD framework, using the intermediate layer ASR at equilibrium and the activations of internal layers of the ``teacher'' BERT models.
In order to make the learning faster and more efficient, we propose transferring knowledge from a pre-trained LM, such as BERT, to our spiking LM. KD is done over the internal layers such as transformer layers, embedding-layer as well as the prediction layer.

To perform transformer layer-based distillation, we utilize the output layer of each SE layer (Fig. \ref{KD_fig}). Subsequently, we establish the loss function by comparing the ASR (at equilibrium) of the output from each SE layer with the activations of the corresponding mapped encoder layers in the ``teacher" model. The loss function using Mean Squared Error (MSE) is formulated as follows,
\begin{equation}
\begin{aligned}
\label{eqn10}
\mathcal{L}_{h_i} = MSE(a_{h_i}^*W_{T_d}, T_{f(h_i)})
\end{aligned}
\end{equation}
where, $a_{h_i}^*$ is the ASR (at equilibrium) of the output neurons of the $i^{th}$ SE layer in the ``student'' and $T_{f(h_i)}$ is the output of the $f(h_i)^{th}$ layer of the ``teacher''. $W_{T_d}$ is a linear transformation that maps the ``student'' layer to the same dimension as the corresponding ``teacher'' network layer. Function $f$ maps ``student'' layer $h_i$ to a specific target layer in the ``teacher'' network. In our approach, we have used the following mapping function: $f(h_i) = h^{'}_{p*i}$; $p = T_{enc}/S_{enc}$, where $h^{'}_i$ is the output of the $i^{th}$ encoder layer of the ``teacher''  and $T_{enc}$ and $S_{enc}$ are the number of encoder layers in the ``teacher'' and ``student'' models respectively.  

By leveraging the equilibrium state of the neurons, KD utilizes the converged ASR to its advantage. Consequently, we employ implicit differentiation technique (Eqn. \ref{eqn3}) for training using the equilibrium state of the intermediate layers. This enables us to perform a faster and efficient layer-wise knowledge transfer from a pre-trained ANN-based LM to a smaller (in size) spiking LM. 
Moreover, each spiking encoder layer within the ``student'' model incorporates a spiking attention layer as described in Fig. \ref{fig1}. We strengthen knowledge transfer further by optimizing an $MSE$ loss over the attention score at equilibrium of the ``student'' network with the attention score of the corresponding mapped attention layer of the ``teacher'' network (following the function $f$). We also perform embedding layer level distillation by formulating a loss function similar to Eqn. \ref{eqn10}. We use ASR of the embedding layer (input to the first spiking encoder layer) and create an MSE loss against the embedding layer of the ``teacher''.

Post-transformer layer distillation, we also perform prediction layer distillation, following the works of \cite{hinton2015distilling}. The loss at the prediction layer for classification tasks can be written as,
\begin{equation}
\begin{aligned}
\label{eqn11}
\mathcal{L}_{pred} = CE(c(a_{pred}^*)/t^{'}, T_{pred}/t^{'})
\end{aligned}
\end{equation}
where, $a_{pred}^*$ is the ASR at equilibrium of the output of the final Spiking Encoder, $c$ acts as a linear mapping and $T_{pred}$ is the output logits of the ``teacher" network. $CE$ is the cross-entropy loss function and $t^{'}$ is temperature.

Distillation is done in two different stages following \cite{jiao2019tinybert}. Firstly, we perform general distillation where we use a pre-trained general BERT model and use general domain data not specific to any particular task. Secondly, we perform task specific distillation on datasets relevant to the particular task using a task-specific fine-tuned BERT model as a ``teacher''. By employing a two-staged distillation process, we significantly enhance the efficiency of our spiking LM development, while also resulting in a substantially reduced ``student'' model size when compared to the ``teacher''.

\section{Experimentation}
In this section, we demonstrate the performance of our proposed spiking LM and evaluate it against different tasks in the General Language Understanding Evaluation (GLUE) benchmark \cite{wang2018glue}. We also highlight the core hyper-parameters for training SpikingBERT model. We also perform extensive analysis to report energy and power efficiency of our proposed model. The experiments were run on Nvidia RTX A5000 GPUs (8) each with 24GB memory.

\subsection{Datasets}
In order to evaluate our model, we chose seven different type of tasks (\textbf{six classification and one regression task}) from the GLUE benchmark. We chose Quora Question Pair (QQP), Microsoft Research Paraphrase Corpus (MRPC) and Semantic Textual Similarity Benchmark (STS-B) (\textbf{regression} task) to evaluate our model on similarity and paraphrase tasks. For inference tasks, we opted for Multi-Genre Natural Language Inference (MNLI), Question-answering NLI (QNLI) and Recognizing Textual Entailment (RTE) datasets. For single-sentence based sentiment analysis tasks, we chose Stanford Sentiment Treebank (SST-2). Further dataset details are available in Appendix A.

\subsection{Baselines \& SpikingBERT Settings}
To the best our knowledge, our proposed model is the first one to report and analyze the performance of a spiking LM on different tasks from the GLUE benchmark. \cite{zhu2023spikegpt} proposed a spiking based GPT architecture and it reported $80.39\%$ accuracy on SST-2 dataset with a model (45M) comparable with our model size (50M). Using a larger model of size 216M, SpikeGPT  achieves $88.76\%$, which is comparable to our performance. Our model is able to demonstrate higher accuracy with lower number of parameters primarily because of the KD technique used to train it as well as because of the BERT-based architecture which is suitable for classification problems due to its bidirectional context understanding. In addition to the above mentioned work, we focus primarily on comparing the performance of our architecture against existing non-spiking methodologies that aims to reduce the complexity of base-BERT model. However, unlike the proposed model, these methods are not applicable for a spiking implementation on neuromorphic chips such as Loihi 2 since all of them are non-spiking architectures. Our goal for the comparisons is to show that low-powered Spiking LM with less trainable parameters can achieve similar accuracy compared to other efficient LM implementations (number of parameters less than 50M). Moreover, since this is the first time spiking LMs have been evaluated against a benchmark, we did not use additional techniques such as data augmentation \cite{jiao2019tinybert}, etc. to boost model performance in order to delineate the core advantages of our proposed training method.

\begin{table}[h!]
    \centering
    \small
\begin{tabular}{ l | l | r } 
 \hline  
 Hyper-parameters &  Range & Optimal \\ 
 \hline
  $T_{conv}$: General KD  & (5-150) & 80\\ 
  $T_{conv}$: Task-based IKD  & (5-150) & 80\\ 
  $V_{th}$ (Threshold Voltage) & (0.25 - 5.0) & 1.0 \\
  $\gamma$ (Leak term) & (0.8 - 1.0) & .99 (LIF); 1 (IF) \\
  $t^{'}$ (Temperature) & (0.1 - 10.0) & 1.0 \\
  Batch Size: General KD & (8-256) &  128 \\
  Batch Size: Task-based IKD & (8-128) &  [16,32] \\
  Epochs: General KD & - &  5 \\
  Epochs: Task-based IKD & - &  20 \\
  \hline
\end{tabular}
\caption{Hyper-parameters (explored range and optimal values) for SpikingBERT\textsubscript{4} used across all datasets.}
\label{hyper_param_table}
\end{table}

For all the tasks, we keep the maximum sequence length at 128. The encoding dimension of the tokens in the input is 768 and the intermediate (IL-2) size of the model is 3072. In order to emphasize on the benefits of KD, SpikingBERT\textsubscript{4} comprises of only 4 SE blocks compared to 12 encoders blocks of BERT. Increasing the number of SE blocks will also improve the overall model performance (analysis in Appendix C). The model trained for reporting the results (Table \ref{table1}) did not have a feedback, since adding it did not increase accuracy.

During training of SpikingBERT, we perform general distillation first using English Wikipedia as text corpus and keeping the sequence length at 128. Transformer (\& Embedding) layer distillation following Eqn. \ref{eqn10} is performed and the ``teacher'' is a pre-trained BERT\textsubscript{BASE} model (uncased). Following this, we perform task-based internal layer KD (IKD) with corresponding fine-tuned BERT models and perform it both on the inner transformer layers and the embedding layer. Core hyper-parameters associated are given in Table \ref{hyper_param_table}. We found that grouping IKD based on type of task (i.e., inference, similarity, etc.) at this stage improves performance. For example, once a task-specific distillation is done using MNLI dataset, if we use that distilled model (as ``student'') and then perform task-specific distillation on QNLI dataset, we achieve higher accuracy on QNLI dataset. After task-based IKD is done, we finally perform prediction-layer distillation following Eqn. \ref{eqn11} to develop the final model. It is to be noted that if we directly train our model on true labels at this stage, we obtain similar results in terms of accuracy. Without using the proposed KD, there is at least 4\% to 5\% drop in accuracy across all datasets. 

\subsection{Analysis of Power \& Energy Efficiency}

\begin{figure}
  \centering
  \includegraphics[width=\columnwidth]{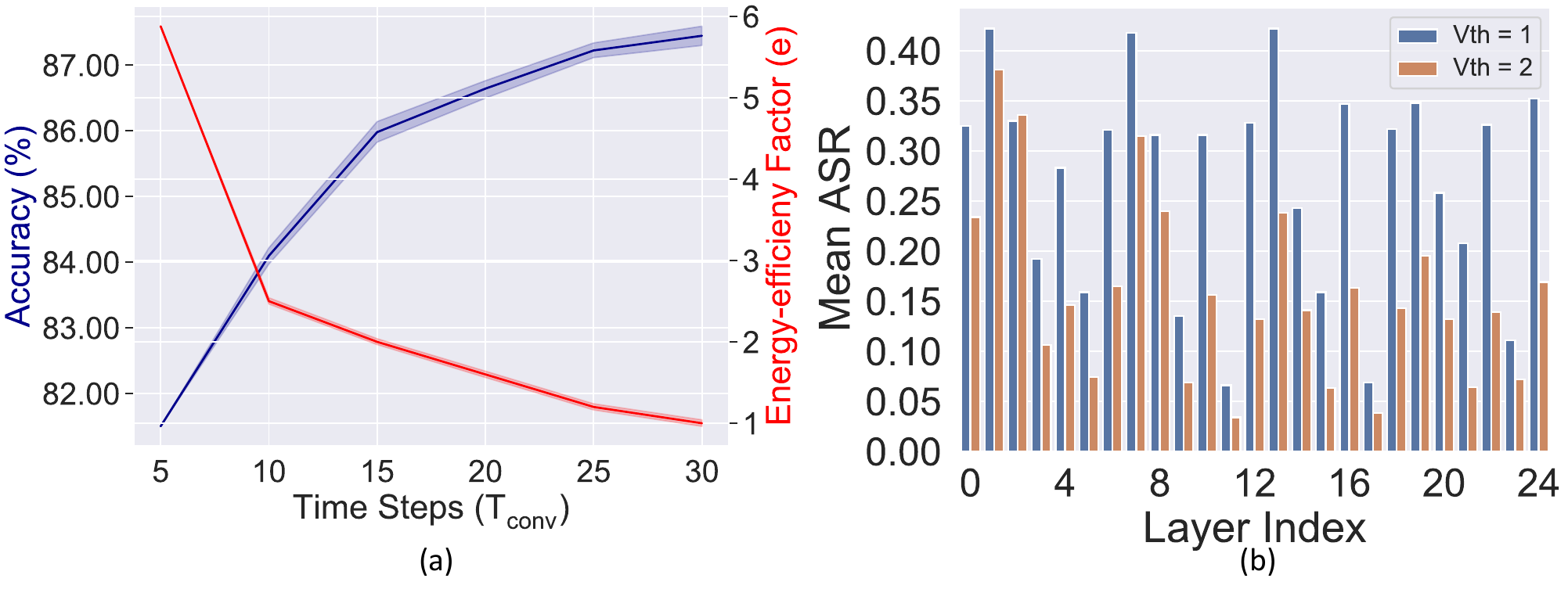}
  \caption{Results obtained on SST-2 dataset.  (a) Variation of Accuracy and Energy-efficiency factor ($e$) as $T_{conv}$ increases. (b) Variation in mean ASR per neuron in different sub-layers of SpikingBERT\textsubscript{4} following changes in $V_{th}$.}
  \label{energy_fig}
\end{figure}

The proposed spiking LM consists of less number of parameters than the ``teacher'' BERT models (109M) and in addition to that it uses only accumulative (ACC) operations in place of multiplicative and accumulative operations (MAC) found in vanilla BERT models. Considering 45nm CMOS technology, ACC operations exhibit an impressive energy efficiency, consuming only 0.9pJ, which is over five times (5.1) more efficient than MAC operations that demand 4.6pJ \cite{han2015learning}.  For estimating the energy and power efficiency of our spiking LM, we leverage the concept of normalized operations \cite{lu2020exploring}, which considers the spiking rates of each layer and corresponding layer-wise operations. The total normalized OPS can be defined as $Norm \#OPS = \frac{\sum_i{IFR_i*Layer\#OPS_{i+1}}}{\sum{Layer\#OPS}}$, where $IFR_i$ is the total number of spikes over inference time steps averaged over number of neurons. Thus, energy-efficiency factor of an SNN, which can be given by the ratio of energy consumed by an iso-architecture ANN over the proposed SNN can be expressed as: $ e = (\frac{1}{5.1} * Norm \#OPS)^{-1}$. SNNs operate on specific time steps, allowing them to dynamically balance accuracy and energy consumption. We perform an extensive energy-accuracy tradeoff analysis on the SST-2 dataset. After conducting general and task-based IKD, during the final training phase, we train a set of models with different values of $T_{conv}$ to see its effect on energy consumption and accuracy. The energy-efficiency factor ($e$) and the obtained accuracy w.r.t the time steps ($T_{conv}$) is demonstrated in Fig. \ref{energy_fig}a. A suitable tradeoff point can be found at $T_{conv} = 16$, where we achieve an accuracy close to the highest value ($2\%$ difference) but we are able to achieve nearly twice energy-efficiency than a non-spiking model of same size.

Moreover, by increasing $V_{th}$, we can reduce the ASR of neurons at each layer, leading to a significant decrease in power consumption. We perform an ablation study on the effects of $V_{th}$ on ASR, which is reported in Fig. \ref{energy_fig}b. Increasing $V_{th}$ intuitively also increases the convergence time steps, thus making energy-consumption effectively similar. However, it allows us to reduce the instantaneous power-consumption considerably - ideal for edge computing. 

\section{Conclusion and Future Works}

%By harnessing bio-plausible neural architectures to design language models, we unlock the potential to establish meaningful connections between learning processes in the human brain and those within LMs. 
Drawing inspiration from the astonishing intricacy of the human brain, a complexity that outshines that of any current LLM, we have the opportunity to leverage these insights in crafting models that not only replicate biologically plausible behavior but also offer energy-efficient solutions through minimal power consumption. In this paper, we propose a spiking LM and evaluate it against multiple tasks in the GLUE benchmark. Leveraging steady-state convergence, we introduced a spiking attention mechanism, proposed a novel ANN-SNN based KD for faster and efficient learning and explored training of Spiking LMs using implicit differentiation, thereby overcoming multiple issues affecting training of SNN models. Implementing our model on neuromorphic hardware such as Loihi 2 for inference will help us develop a low-powered solution which can potentially be implemented on edge devices. 

Further endeavours can be made to extend this methodology to design other spiking LMs such as GPT, etc.  There is still a performance gap between the proposed spiking LM and BERT-based fine-tuned models. We can work towards closing this gap by delving into diverse spiking neuron models, examining temporal encoding schemes, and incorporating graded spikes, among other strategies.

\section{{Acknowledgments}}
The work was supported by the National Science Foundation CAREER Award under Grant \#2337646 and by Oracle Cloud credits and related resources provided by the Oracle for Research program.

\clearpage
\renewcommand{\thefigure}{A\arabic{figure}}

\setcounter{figure}{0}
\setcounter{equation}{0}

\section{Appendix}
\label{sec:appendix}

\subsection{Appendix A: Dataset Details}
The GLUE benchmark serves as a standardized testbed designed to assess the proficiency of a language model's performance. We  selected seven distinct tasks from the GLUE benchmark, aiming to comprehensively assess our model across a set of diverse NLP tasks. These tasks encompass a range of linguistic aspects, including similarity and paraphrasing, natural language inference, and sentiment analysis—representing the three overarching categories that constitute the GLUE benchmark. Among the seven chosen tasks, six pertain to text classification, while the remaining one involves a regression task. The selected tasks and their corresponding datasets are briefly discussed below.

\subsubsection{Quora Question Pairs(QQP)}
This dataset is a collection of question pairs from the QA-website Quora and the task involved is to determine whether two questions are semantically similar or not. 

\subsection{Microsoft Research Paraphrase Corpus (MRPC)}
This dataset consists of sentence pairs extracted from online news sources along with human annotated labels. The task is to determine whether the sentences in the pair are semantically equivalent or not.

\subsection{Semantic Textual Similarity Benchmark (STS-B)}
This dataset consists of sentence pairs from different sources such as video and image captions, NLI data, etc. The task is to assign a score between 1-5 to determine how similar the two sentences are. STS-B is a regression task.

\subsection{Multi-Genre Natural Language Inference (MNLI)}
This dataset consists of a crowd-sourced collection of sentence pairs consisting of a premise and a hypothesis. The task is to determine whether the premise entails or contradicts the hypothesis or whether the relationship is neutral.

\subsection{Question-answering NLI (QNLI)}
This dataset consists of question-paragraph pairs (from Wikipedia). The task is to determine whether the answer to the question is present in the given paragraph.

\subsection{Recognizing Textual Entailment (RTE)}
This dataset consists of data from annual textual entailment challenges. Data created is based on news and Wikipedia. The task is to predict whether one sentence is entailed from the other sentence or not.

\subsection{Stanford Sentiment Treebank (SST-2)}
This dataset consists of specific sentences from movie reviews and a label associated with them stating whether it is positive or negative. Objective is to predict whether the sentiment of the sentence is positive or negative.

\subsection{Appendix B: Additional Architectural Details}

In this section, we explain the exact operations inside the intermediate layers (IL-1, IL-2) and output layers.
The  membrane potential of LIF neurons in IL-1 sub-layer is given as,
\begin{equation}
\begin{aligned}
\label{eqnappn1}
u_{(IL1)}[t+1] = \gamma u_{(IL1)}[t] + norm(W_{(IL1)}s_{attn}[t+1] \\+ s_{input}[t+1])   + b_{(IL1)}  - V_{th}s_{IL1}[t+1]
\end{aligned}
\end{equation}
where, $W_{(IL1)}$ is a linear mapping, $s_{attn}[t]$ are the spikes fed from the spiking attention layer, $s_{input}$ are the spikes from the input layer and $b_{(IL1)}$ is the bias. The  membrane potential of LIF neurons in the IL-2 sub-layer is given as,
\begin{equation}
\begin{aligned}
\label{eqnappn2}
u_{(IL2)}[t+1] = \gamma u_{(IL2)}[t] + gelu(W_{(IL2)}s_{IL1}[t+1])   \\ + b_{(IL2)}  - V_{th}s_{IL2}[t+1]
\end{aligned}
\end{equation}
where, $gelu$ is an activation function (approximately linear in the range [0,1]), $W_{(IL2)}$ is a linear mapping, $s_{IL1}[t]$ are the spikes fed from the IL-1 layer and $b_{(IL2)}$ is the bias. The membrane potential of LIF neurons in the output layer is similar to Eqn. \ref{eqnappn1}, but the input are the spikes from the IL-2 layer.

Surrogate steady-state function at equilibrium for IL-1 can be formulated as, $a_{(IL1)}^* = \sigma (\frac{1}{V_{th}}(norm(W_{(IL1)}a^*_{attn}+ a^*_{input}) + b_{(IL1)}))$, for IL-2 as $a_{(IL2)}^* = \sigma (\frac{1}{V_{th}}(gelu(W_{(IL2)}a^*_{IL1})   + b_{(IL2)}))$ and for output layer of each SE layer as $a_{(output)}^* = \sigma (\frac{1}{V_{th}}(norm(W_{(output)}a^*_{IL2}+ a^*_{IL1})   + b_{(output)}))$.

\subsection{Appendix C: Additional Experimental Details}

In this subsection, we briefly go over the implementation details and additional ablation studies performed to analyse our model performance. The GPU used for the experiments are already given in the main paper and the CPU used is Intel(R) Xeon(R) Gold 6226R CPU @ 2.90GHz (64).

\subsubsection{Implementation Details}
We have primarily used Python for writing the code of the described methodology. The pre-trained models for BERT\textsubscript{BASE} and other task-based fine-tuned BERT models were taken from Huggingface. For updating the weights, we used BertAdam, a modified variant of Adam optimizer with settings: $b1=0.9, b2=0.999, ep = 1e-6$. Learning rate is kept at 4e-5 for general KD and 2e-5 for task-based KD as well as for prediction layer distillation. After successful general KD and task-based IKD, only 2 to 4 epochs are enough for prediction-layer KD. General KD takes around 1 day on the GPU specified in the paper (given $T_{conv}=125$). Task-based IKD depends on $T_{conv}$ as well as the size of the dataset (est. 8 hours to a day). Prediction-layer distillation in general is a lot faster since accuracy converges within 2 to 4 epochs.
\begin{figure}
  \centering
  \includegraphics[width=\columnwidth]{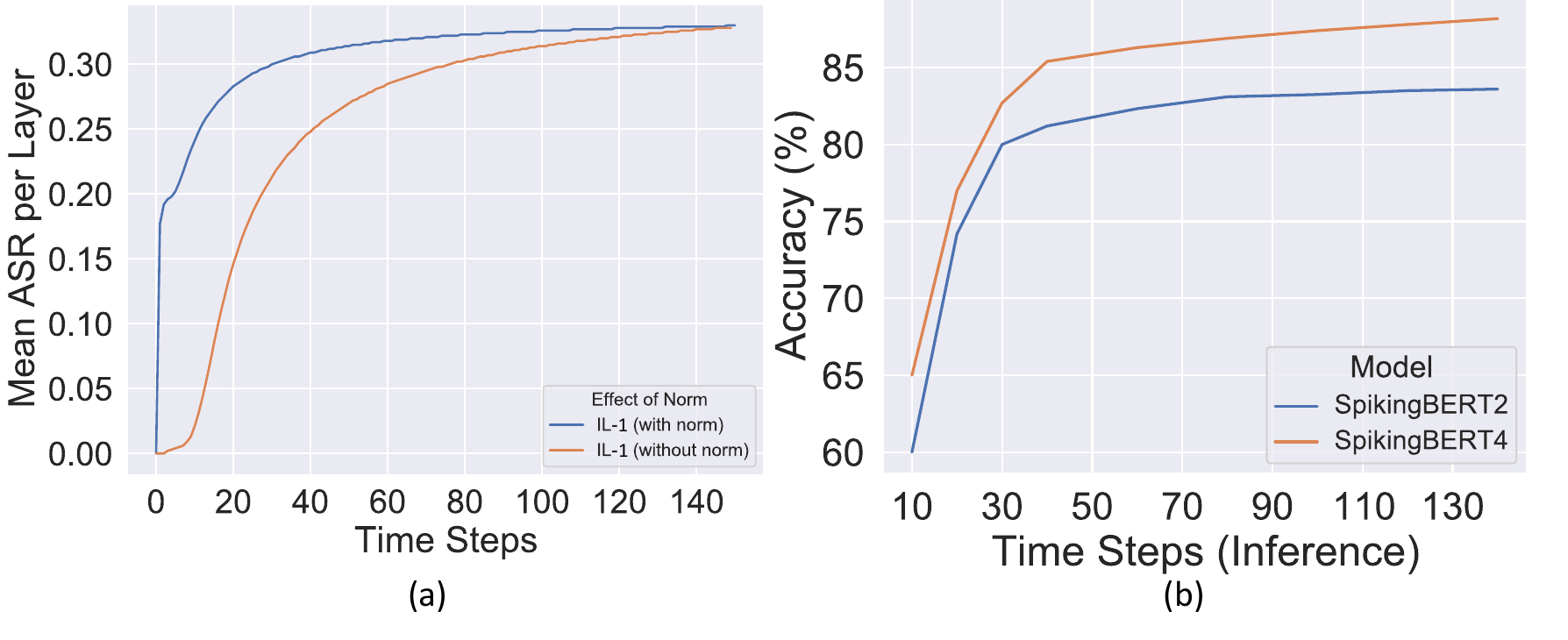}
  \caption{ \small (a) Graph illustrating mean (over number of neurons) of the ASR of the IL-1 sub-layer (in an SE layer) against the operating time steps. This visualization contrasts scenarios where the final model employs normalization with those where normalization is omitted. (b) Graph showing accuracy against time steps used for convergence during inference for SpikingBERT models with 2 and 4 SE layers.}
  \label{il1_conv}
\end{figure}

\subsubsection{Ablation Studies}

We performed a series of  ablation studies to analyse the behavior of our model.

\noindent \textbf{(a) Effect of normalization:} For increased bio-plausibility, we trained a network without using layer normalization in Eqn. \ref{eqnappn1}, thereby making the operation within those neurons completely linear. For the experiment performed on SST-2 dataset, we first did 5 epochs of general KD and 10 epochs of task-based IKD (SST-2 specific) with the usual model architecture (with normalization). Then we performed 10 epochs of task-based IKD (SST-2 specific) by removing the layer normalization layer from IL-1 and output layers (since there is no normalization in IL-2 and spiking attention). The resulting final model (without normalization) performed similar to the model with normalization. However, the time steps for convergence increased considerably as shown in Fig \ref{il1_conv}(a). Though this is a preliminary observation, future explorations can be made to understand the effects of normalization in spiking architectures. One further observation we noted during the experiments was that, if a model was developed from scratch without normalization (i.e. without the additional steps done in the above experiment like GD and task-based IKD with normalization), then there was around 4\% drop in accuracy (in SST-2 dataset) with the convergence behavior still remaining similar.

\noindent \textbf{(b) Effect of increase in SE layers:} 
In the paper, we used 4 SE layers in our spiking LM. Increasing the number of SE layers results in increase of the model performance. We performed an experiment where we trained two different models, viz. SpikingBERT\textsubscript{2} (2 SE layers) and SpikingBERT\textsubscript{4} (4 SE layers). After completing both general KD and task-based IKD (SST-2 dataset) for the two models, we did prediction layer distillation on both networks for $T_{conv} = 140$. It is to be noted that unlike in the energy analysis study done in the main paper where we trained a set of models (with prediction-layer distillation) for different values of $T_{conv}$, in this experiment we trained the networks using only $T_{conv}=140$. In the graph demonstrated in Fig. \ref{il1_conv}(b), we show accuracy w.r.t different values of convergence time steps ($T_{conv}$) used during inference. Our focus in this paper was to delineate an efficient framework capable of training a deep spiking LM. Future research can capitalize on the framework and focus on developing architectures with more number of layers (greater than four spiking encoder layers).


\begin{thebibliography}{35}
\providecommand{\natexlab}[1]{#1}

\bibitem[{Alawad, Yoon, and Tourassi(2017)}]{alawad2017energy}
Alawad, M.; Yoon, H.-J.; and Tourassi, G. 2017.
\newblock Energy efficient stochastic-based deep spiking neural networks for sparse datasets.
\newblock In \emph{2017 IEEE International Conference on Big Data (Big Data)}, 311--318. IEEE.

\bibitem[{Amir et~al.(2017)Amir, Taba, Berg, Melano, McKinstry, Di~Nolfo, Nayak, Andreopoulos, Garreau, Mendoza et~al.}]{amir2017low}
Amir, A.; Taba, B.; Berg, D.; Melano, T.; McKinstry, J.; Di~Nolfo, C.; Nayak, T.; Andreopoulos, A.; Garreau, G.; Mendoza, M.; et~al. 2017.
\newblock A low power, fully event-based gesture recognition system.
\newblock In \emph{Proceedings of the IEEE conference on computer vision and pattern recognition}, 7243--7252.

\bibitem[{Bai, Kolter, and Koltun(2019)}]{bai2019deep}
Bai, S.; Kolter, J.~Z.; and Koltun, V. 2019.
\newblock Deep equilibrium models.
\newblock \emph{Advances in Neural Information Processing Systems}, 32.

\bibitem[{Bal and Sengupta(2022)}]{bal2022sequence}
Bal, M.; and Sengupta, A. 2022.
\newblock Sequence Learning using Equilibrium Propagation.
\newblock \emph{arXiv preprint arXiv:2209.09626}.

\bibitem[{Brown et~al.(2020)Brown, Mann, Ryder, Subbiah, Kaplan, Dhariwal, Neelakantan, Shyam, Sastry, Askell et~al.}]{brown2020language}
Brown, T.; Mann, B.; Ryder, N.; Subbiah, M.; Kaplan, J.~D.; Dhariwal, P.; Neelakantan, A.; Shyam, P.; Sastry, G.; Askell, A.; et~al. 2020.
\newblock Language models are few-shot learners.
\newblock \emph{Advances in neural information processing systems}, 33: 1877--1901.

\bibitem[{Davies et~al.(2021)Davies, Wild, Orchard, Sandamirskaya, Guerra, Joshi, Plank, and Risbud}]{davies2021advancing}
Davies, M.; Wild, A.; Orchard, G.; Sandamirskaya, Y.; Guerra, G. A.~F.; Joshi, P.; Plank, P.; and Risbud, S.~R. 2021.
\newblock Advancing neuromorphic computing with loihi: {A} survey of results and outlook.
\newblock \emph{Proceedings of the IEEE}, 109(5): 911--934.

\bibitem[{Deng et~al.(2009)Deng, Dong, Socher, Li, Li, and Fei-Fei}]{deng2009imagenet}
Deng, J.; Dong, W.; Socher, R.; Li, L.-J.; Li, K.; and Fei-Fei, L. 2009.
\newblock Imagenet: A large-scale hierarchical image database.
\newblock In \emph{2009 IEEE conference on computer vision and pattern recognition}, 248--255. Ieee.

\bibitem[{Devlin et~al.(2018)Devlin, Chang, Lee, and Toutanova}]{devlin2018bert}
Devlin, J.; Chang, M.-W.; Lee, K.; and Toutanova, K. 2018.
\newblock Bert: Pre-training of deep bidirectional transformers for language understanding.
\newblock \emph{arXiv preprint arXiv:1810.04805}.

\bibitem[{Frantar, Kurtic, and Alistarh(2021)}]{frantar2021m}
Frantar, E.; Kurtic, E.; and Alistarh, D. 2021.
\newblock M-FAC: Efficient matrix-free approximations of second-order information.
\newblock \emph{Advances in Neural Information Processing Systems}, 34: 14873--14886.

\bibitem[{Ghosh-Dastidar and Adeli(2009)}]{ghosh2009spiking}
Ghosh-Dastidar, S.; and Adeli, H. 2009.
\newblock Spiking neural networks.
\newblock \emph{International journal of neural systems}, 19(04): 295--308.

\bibitem[{Han et~al.(2015)Han, Pool, Tran, and Dally}]{han2015learning}
Han, S.; Pool, J.; Tran, J.; and Dally, W.~J. 2015.
\newblock Learning both Weights and Connections for Efficient Neural Networks.
\newblock arXiv:1506.02626.

\bibitem[{Hinton, Vinyals, and Dean(2015)}]{hinton2015distilling}
Hinton, G.; Vinyals, O.; and Dean, J. 2015.
\newblock Distilling the knowledge in a neural network.
\newblock \emph{arXiv preprint arXiv:1503.02531}.

\bibitem[{Hong et~al.(2023)Hong, Shen, Qi, and Wang}]{hong2023lasnn}
Hong, D.; Shen, J.; Qi, Y.; and Wang, Y. 2023.
\newblock LaSNN: Layer-wise ANN-to-SNN Distillation for Effective and Efficient Training in Deep Spiking Neural Networks.
\newblock \emph{arXiv preprint arXiv:2304.09101}.

\bibitem[{Jiao et~al.(2019)Jiao, Yin, Shang, Jiang, Chen, Li, Wang, and Liu}]{jiao2019tinybert}
Jiao, X.; Yin, Y.; Shang, L.; Jiang, X.; Chen, X.; Li, L.; Wang, F.; and Liu, Q. 2019.
\newblock Tinybert: Distilling bert for natural language understanding.
\newblock \emph{arXiv preprint arXiv:1909.10351}.

\bibitem[{Kim et~al.(2021)Kim, Gholami, Yao, Mahoney, and Keutzer}]{kim2021bert}
Kim, S.; Gholami, A.; Yao, Z.; Mahoney, M.~W.; and Keutzer, K. 2021.
\newblock I-bert: Integer-only bert quantization.
\newblock In \emph{International conference on machine learning}, 5506--5518. PMLR.

\bibitem[{Krizhevsky, Nair, and Hinton(2009)}]{krizhevsky2009cifar}
Krizhevsky, A.; Nair, V.; and Hinton, G. 2009.
\newblock Cifar-10 and cifar-100 datasets.
\newblock \emph{URl: https://www. cs. toronto. edu/kriz/cifar. html}, 6(1): 1.

\bibitem[{Kubilius et~al.(2019)Kubilius, Schrimpf, Kar, Rajalingham, Hong, Majaj, Issa, Bashivan, Prescott-Roy, Schmidt et~al.}]{kubilius2019brain}
Kubilius, J.; Schrimpf, M.; Kar, K.; Rajalingham, R.; Hong, H.; Majaj, N.; Issa, E.; Bashivan, P.; Prescott-Roy, J.; Schmidt, K.; et~al. 2019.
\newblock Brain-like object recognition with high-performing shallow recurrent ANNs.
\newblock \emph{Advances in neural information processing systems}, 32.

\bibitem[{Kurtic et~al.(2022)Kurtic, Campos, Nguyen, Frantar, Kurtz, Fineran, Goin, and Alistarh}]{kurtic2022optimal}
Kurtic, E.; Campos, D.; Nguyen, T.; Frantar, E.; Kurtz, M.; Fineran, B.; Goin, M.; and Alistarh, D. 2022.
\newblock The optimal bert surgeon: Scalable and accurate second-order pruning for large language models.
\newblock \emph{arXiv preprint arXiv:2203.07259}.

\bibitem[{Lee et~al.(2020)Lee, Sarwar, Panda, Srinivasan, and Roy}]{lee2020enabling}
Lee, C.; Sarwar, S.~S.; Panda, P.; Srinivasan, G.; and Roy, K. 2020.
\newblock Enabling spike-based backpropagation for training deep neural network architectures.
\newblock \emph{Frontiers in neuroscience}, 119.

\bibitem[{Lu and Sengupta(2020)}]{lu2020exploring}
Lu, S.; and Sengupta, A. 2020.
\newblock Exploring the connection between binary and spiking neural networks.
\newblock \emph{Frontiers in neuroscience}, 14: 535.

\bibitem[{Neftci, Mostafa, and Zenke(2019)}]{neftci2019surrogate}
Neftci, E.~O.; Mostafa, H.; and Zenke, F. 2019.
\newblock Surrogate gradient learning in spiking neural networks: Bringing the power of gradient-based optimization to spiking neural networks.
\newblock \emph{IEEE Signal Processing Magazine}, 36(6): 51--63.

\bibitem[{Orchard et~al.(2015)Orchard, Jayawant, Cohen, and Thakor}]{orchard2015converting}
Orchard, G.; Jayawant, A.; Cohen, G.~K.; and Thakor, N. 2015.
\newblock Converting static image datasets to spiking neuromorphic datasets using saccades.
\newblock \emph{Frontiers in neuroscience}, 9: 437.

\bibitem[{Radford et~al.(2018)Radford, Narasimhan, Salimans, Sutskever et~al.}]{radford2018improving}
Radford, A.; Narasimhan, K.; Salimans, T.; Sutskever, I.; et~al. 2018.
\newblock Improving language understanding by generative pre-training.

\bibitem[{Scellier and Bengio(2017)}]{scellier2017equilibrium}
Scellier, B.; and Bengio, Y. 2017.
\newblock Equilibrium propagation: Bridging the gap between energy-based models and backpropagation.
\newblock \emph{Frontiers in computational neuroscience}, 11: 24.

\bibitem[{Sengupta et~al.(2019)Sengupta, Ye, Wang, Liu, and Roy}]{sengupta2019going}
Sengupta, A.; Ye, Y.; Wang, R.; Liu, C.; and Roy, K. 2019.
\newblock Going deeper in spiking neural networks: VGG and residual architectures.
\newblock \emph{Frontiers in neuroscience}, 13: 95.

\bibitem[{Takuya, Zhang, and Nakashima(2021)}]{takuya2021training}
Takuya, S.; Zhang, R.; and Nakashima, Y. 2021.
\newblock Training low-latency spiking neural network through knowledge distillation.
\newblock In \emph{2021 IEEE Symposium in Low-Power and High-Speed Chips (COOL CHIPS)}, 1--3. IEEE.

\bibitem[{Tang et~al.(2019)Tang, Lu, Liu, Mou, Vechtomova, and Lin}]{tang2019distilling}
Tang, R.; Lu, Y.; Liu, L.; Mou, L.; Vechtomova, O.; and Lin, J. 2019.
\newblock Distilling task-specific knowledge from bert into simple neural networks.
\newblock \emph{arXiv preprint arXiv:1903.12136}.

\bibitem[{Vaswani et~al.(2017)Vaswani, Shazeer, Parmar, Uszkoreit, Jones, Gomez, Kaiser, and Polosukhin}]{vaswani2017attention}
Vaswani, A.; Shazeer, N.; Parmar, N.; Uszkoreit, J.; Jones, L.; Gomez, A.~N.; Kaiser, L.; and Polosukhin, I. 2017.
\newblock Attention Is All You Need.
\newblock arXiv:1706.03762.

\bibitem[{Wang et~al.(2018)Wang, Singh, Michael, Hill, Levy, and Bowman}]{wang2018glue}
Wang, A.; Singh, A.; Michael, J.; Hill, F.; Levy, O.; and Bowman, S.~R. 2018.
\newblock GLUE: A multi-task benchmark and analysis platform for natural language understanding.
\newblock \emph{arXiv preprint arXiv:1804.07461}.

\bibitem[{Wei et~al.(2022)Wei, Tay, Bommasani, Raffel, Zoph, Borgeaud, Yogatama, Bosma, Zhou, Metzler, Chi, Hashimoto, Vinyals, Liang, Dean, and Fedus}]{wei2022emergent}
Wei, J.; Tay, Y.; Bommasani, R.; Raffel, C.; Zoph, B.; Borgeaud, S.; Yogatama, D.; Bosma, M.; Zhou, D.; Metzler, D.; Chi, E.~H.; Hashimoto, T.; Vinyals, O.; Liang, P.; Dean, J.; and Fedus, W. 2022.
\newblock Emergent Abilities of Large Language Models.
\newblock arXiv:2206.07682.

\bibitem[{Xiao et~al.(2021)Xiao, Meng, Zhang, Wang, and Lin}]{xiao2021training}
Xiao, M.; Meng, Q.; Zhang, Z.; Wang, Y.; and Lin, Z. 2021.
\newblock Training feedback spiking neural networks by implicit differentiation on the equilibrium state.
\newblock \emph{Advances in Neural Information Processing Systems}, 34: 14516--14528.

\bibitem[{Xu et~al.(2021)Xu, Tan, Luo, Song, Li, Qin, and Liu}]{xu2021bert}
Xu, J.; Tan, X.; Luo, R.; Song, K.; Li, J.; Qin, T.; and Liu, T.-Y. 2021.
\newblock NAS-BERT: task-agnostic and adaptive-size BERT compression with neural architecture search.
\newblock In \emph{Proceedings of the 27th ACM SIGKDD Conference on Knowledge Discovery \& Data Mining}, 1933--1943.

\bibitem[{Xu et~al.(2023)Xu, Li, Shen, Liu, Tang, and Pan}]{xu2023constructing}
Xu, Q.; Li, Y.; Shen, J.; Liu, J.~K.; Tang, H.; and Pan, G. 2023.
\newblock Constructing deep spiking neural networks from artificial neural networks with knowledge distillation.
\newblock In \emph{Proceedings of the IEEE/CVF Conference on Computer Vision and Pattern Recognition}, 7886--7895.

\bibitem[{Zhou et~al.(2022)Zhou, Zhu, He, Wang, Yan, Tian, and Yuan}]{zhou2022spikformer}
Zhou, Z.; Zhu, Y.; He, C.; Wang, Y.; Yan, S.; Tian, Y.; and Yuan, L. 2022.
\newblock Spikformer: When spiking neural network meets transformer.
\newblock \emph{arXiv preprint arXiv:2209.15425}.

\bibitem[{Zhu, Zhao, and Eshraghian(2023)}]{zhu2023spikegpt}
Zhu, R.-J.; Zhao, Q.; and Eshraghian, J.~K. 2023.
\newblock Spikegpt: Generative pre-trained language model with spiking neural networks.
\newblock \emph{arXiv preprint arXiv:2302.13939}.

\end{thebibliography}
\end{document}